\documentclass[10pt,twocolumn,letterpaper]{article}

\usepackage{cvpr}              

\usepackage[dvipsnames]{xcolor}

\definecolor{cvprblue}{rgb}{0.21,0.49,0.74}
\usepackage[pagebackref,breaklinks,colorlinks,citecolor=cvprblue]{hyperref}

\def\paperID{9} 
\def\confName{CVPR}
\def\confYear{2024}

\title{Transformers meet Neural Algorithmic Reasoners}
\definecolor{mygreen}{rgb}{0,0.6,0}
\author{Wilfried Bounsi\\
Google DeepMind\\
{\tt\small wilcoln@google.com}
\and
Borja Ibarz\\
Google DeepMind\\
{\tt\small bibarz@google.com}
\and
Andrew Dudzik\\
Google DeepMind\\
{\tt\small adudzik@google.com}
\and
Jessica B. Hamrick\\
Google DeepMind\\
{\tt\small jhamrick@google.com}
\and
Larisa Markeeva\\
Google DeepMind\\
{\tt\small lmarkeeva@google.com}
\and
Alex Vitvitskyi\\
Google DeepMind\\
{\tt\small avlife@google.com}
\and
Razvan Pascanu\\
Google DeepMind\\
{\tt\small razp@google.com}
\and
Petar Veli\v{c}kovi\'{c}\\
Google DeepMind\\
{\tt\small petarv@google.com}
}

\begin{document}
\maketitle
\begin{abstract}
Transformers have revolutionized machine learning with their simple yet effective architecture. Pre-training Transformers on massive text datasets from the Internet has led to unmatched generalization for natural language understanding (NLU) tasks. However, such language models remain fragile when tasked with algorithmic forms of reasoning, where computations must be precise and robust. To address this limitation, we propose a novel approach that combines the Transformer's language understanding with the robustness of graph neural network (GNN)-based neural algorithmic reasoners (NARs). Such NARs proved effective as generic solvers for algorithmic tasks, when specified in graph form. To make their embeddings accessible to a Transformer, we propose a hybrid architecture with a two-phase training procedure, allowing the tokens in the language model to cross-attend to the node embeddings from the NAR. We evaluate our resulting TransNAR model on CLRS-Text, the text-based version of the CLRS-30 benchmark, and demonstrate significant gains over Transformer-only models for algorithmic reasoning, both in and out of distribution.
\end{abstract}    
\section{Introduction}
\begin{figure}
    \centering
    \includegraphics[width=\columnwidth]{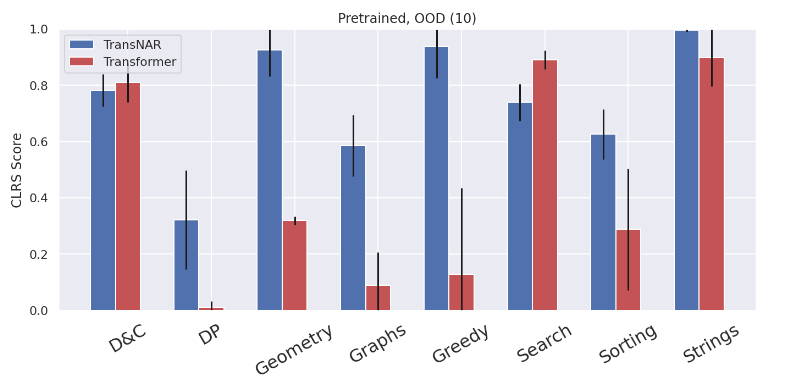}
    \caption{Our TransNAR architecture, with its direct synergy of Transformers and Neural Algorithmic Reasoners, yields clear improvements in out-of-distribution reasoning across wide categories of algorithmic tasks in CLRS-Text \citep{markeeva2024clrstext}, a textual version of the CLRS-30 benchmark \citep{clrs30}. Here, the $x$-axis indicates one of the eight algorithmic families of CLRS-30, and the $y$-axis spans the average execution accuracy across a dataset of out-of-distribution examples. TransNAR enables \emph{emerging capabilities} in the particular out-of-distribution regime depicted here, with over 20\% absolute improvement in several of the algorithm classes.}
    \label{fig:blockplot}
\end{figure}
Recent work motivated \cite{gnnsaredps} and showcased \cite{generalist, causalreg} the effectiveness of graph neural networks \citep[GNNs]{velivckovic2023everything} at robustly solving algorithmic tasks of various input sizes, both in and out of distribution---such systems are often referred to as \emph{neural algorithmic reasoners} \citep[NARs]{nar}. Provided appropriate inductive biases are used, NARs are capable of holding perfect generalisation even on $6\times$ larger inputs than ones seen in the training set, for highly complex algorithmic tasks with long rollouts \citep{jurss2023recursive}. NARs are, however, still relatively \emph{narrow} forms of AI, as they require rigidly structured formatting of inputs, and they hence cannot be directly applied to problems posed in more noisy forms---such as in \emph{natural language}---even when the underlying problem is still algorithmic in nature.

\begin{figure*}[ht]
\begin{center}
\includegraphics[width=0.8\linewidth]{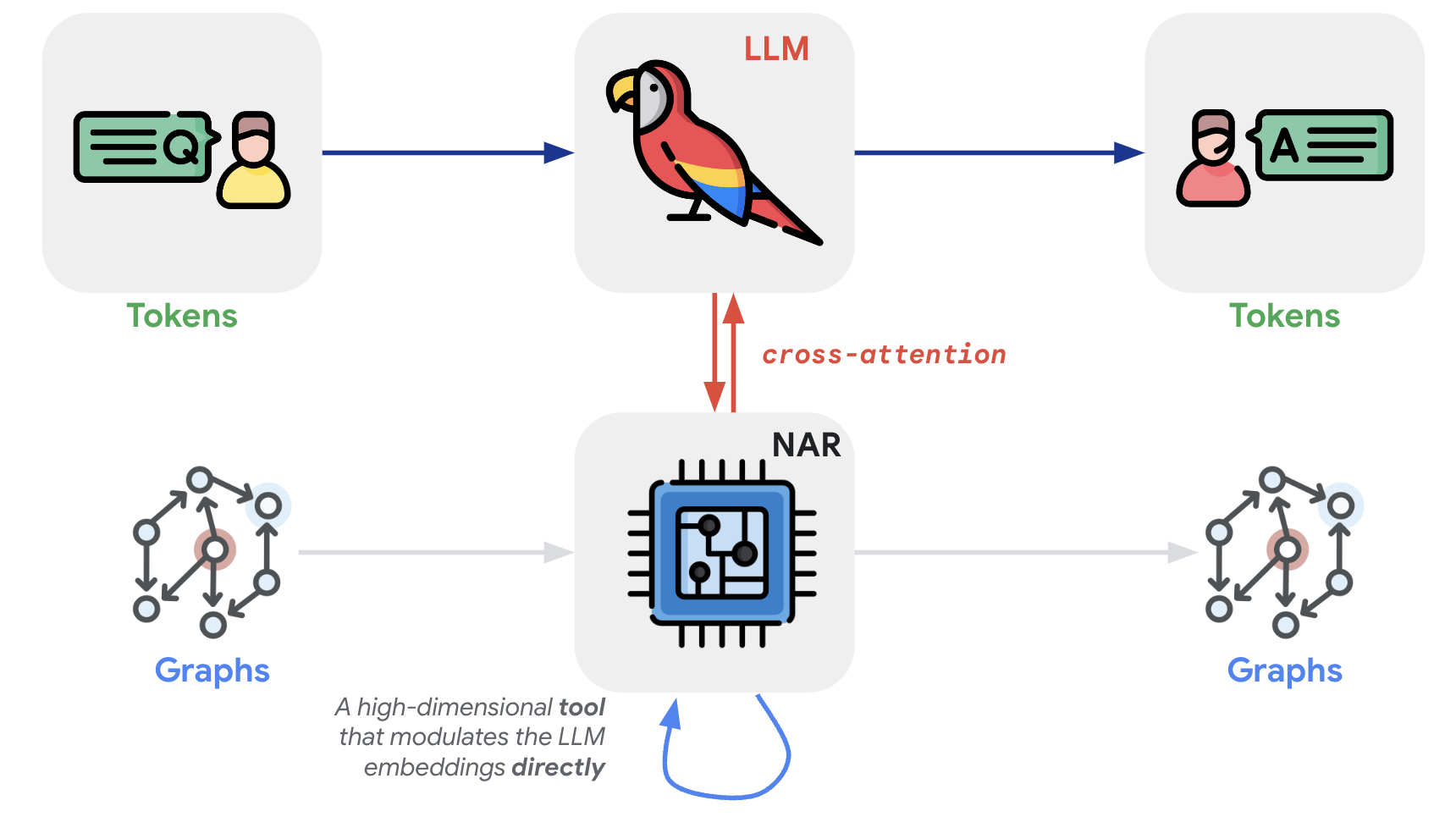}
\caption{{\bf Augmenting LLMs with algorithmic reasoning: a bird's eye view of TransNAR.} A large language model (LLM) consumes input tokens and produces output tokens, as common for a unimodal Transformer. The neural algorithmic reasoner (NAR) module is a graph neural network (GNN) pre-trained to execute various algorithmic computation on a collection of graph-based inputs \citep{generalist}---the pre-training pipeline is denoted by faded arrows. Throughout its forward pass, the Transformer may access the embeddings computed by the NAR, by leveraging cross-attention (trained by learnable ``glue'' weights).}
\label{idea}
\end{center}
\end{figure*}

Conversely, the current undisputed state-of-the-art approach for modelling noisy text data are Transformer-based \citep{vaswani2017attention} language models \citep{team2023gemini,achiam2023gpt}. In spite of their unrivalled natural language understanding properties, they are also notoriously brittle when faced with even the simplest algorithmic tasks \citep{dziri2023faith}---especially if out-of-distribution generalisation is required \citep{anil2022exploring}.

It appears that \emph{\textbf{uniting} Transformers with NARs} can lead to fruitful returns on both sides. In this paper, we explore this interface for the first time, building the \textbf{TransNAR} model.

\paragraph{Contributions} 

Our exploration proved fruitful. The key takeaways we present in this work are as follows:
\begin{enumerate}
    \item We propose a hybrid architecture combining the language understanding of a Transformer with the robustness of reasoning of a pre-trained GNN-based NAR. The Transformer uses the NAR as a \emph{high-dimensional tool} that will modulate its tokens embeddings.
    \item We show, through an evaluation on CLRS-Text \citep{markeeva2024clrstext}, the text-based version of the CLRS-30 benchmark, that such a NAR-augmented large language model (LLM) exhibits improved and more robust reasoning capabilities out-of-distribution (Figure \ref{fig:blockplot}).
\end{enumerate}

Our work presents one of the most comprehensive size generalisation challenges given to Transformers to date, and the introduction of NARs moves the needle significantly.
\section{Related work}

Our work sits at the intersection of several areas: neural algorithmic reasoning, length generalisation in language models, tool use, and multimodality. Here, we briefly survey various relevant works in each area. Due to the diversity of perspectives, to preserve brevity, we do not offer a comprehensive review of related work, but rather aim to provide an indication of specific works that inspired ours the most.

\paragraph{Neural algorithmic reasoning} NAR is, in general terms, the art of building neural networks that are capable of capturing algorithmic computation. Such capabilities can be amplified by careful choices in algorithmic alignment \citep{xu2020neural}, step-wise training \citep{velivckovic2019neural} or contrastive objectives \citep{causalreg}. 

Recently, it was demonstrated that: (1) it is possible to learn an NAR capable of executing \emph{multiple} algorithms simultaneously in its latent space \citep{xhonneux2021transfer}---with the Triplet-GMPNN \citep{generalist} skillfully doing so for a collection of thirty algorithms across the CLRS benchmark \citep{clrs30}; (2) Once trained, such NARs can be usefully deployed in various downstream tasks: reinforcement learning \citep{deac2021neural,he2022continuous}, self-supervised learning \citep{velivckovic2022reasoning}, combinatorial optimisation \citep{georgiev2023neural,qian2023exploring}, computational biology \citep{georgievnarti} and neuroscience \citep{numeroso2023dual}.

Our work's use of NAR is mostly motivated by two of the works listed before: we use a relatively small, pre-trained, multi-task NAR \citep{generalist}, and deploy it in a far more scaled environment: as shown by \citet{numeroso2023dual}, NAR should in principle be scalable to systems that are orders-of-magnitude greater than the NAR's training distribution ($180,000\times$ in that particular case).

\begin{figure*}[ht]
\begin{center}
\includegraphics[width=\linewidth]{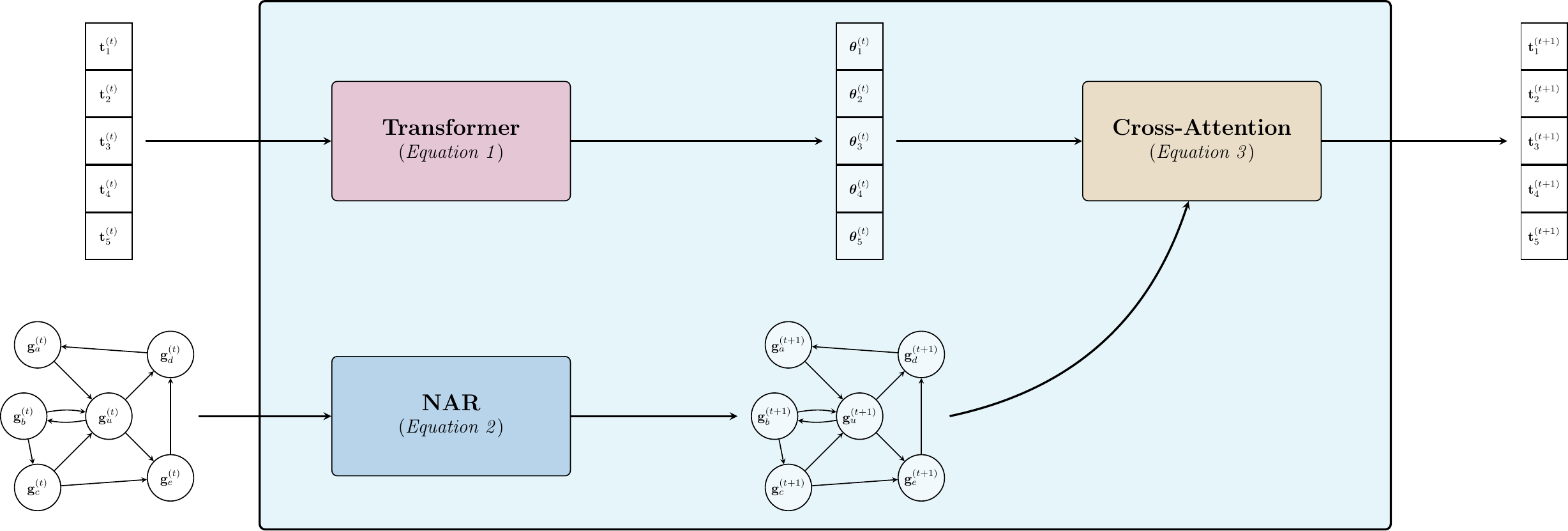}
\caption{\textbf{TransNAR hybrid architecture.} Similar to \citet{flamingo}, we interleave existing Transformer layers with gated cross-attention layers which enable information to flow from the NAR to the Transformer. We generate queries from tokens while we obtain keys and values from nodes and edges of the graph. The node and edge embeddings are obtained by running the NAR on the graph version of the reasoning task to be solved. When experimenting with pre-trained Transformers, we initially close the cross-attention gate, in order to fully preserve the language model's internal knowledge at the beginning of training.}
\label{fig:architecture}
\end{center}
\end{figure*}

\paragraph{Length generalisation in LLMs}

While NARs can often strongly generalise to far greater test inputs \citep{jurss2023recursive}, LLMs have seen significantly less success in such scenarios. We attribute this to their autoregressive, causally-masked objective, which may not always correspond to the most logical order in which outputs of algorithms should be predicted. Just as a simple example, performance of various LLMs on multiplication can be significantly improved by predicting the result in reverse order \citep{lee2023teaching}. Of course, on more complicated algorithms, it may be much harder to determine the best way to permute the input, and it may not be the most human-readable. 

Knowledge of the above issues has led to a significant amount of effort being invested in building Transformers that can generalise in length. While length generalisation is not the only kind of distribution shift of interest to OOD reasoning, it is among the most easy such shifts to simulate. Accordingly, various works have attempted to induce length generalisation in LLMs, through the use of careful prompting \citep{zhou2022teaching,shen2023positional}, randomised positional encoding \citep{randomizedpe}, curricula \citep{abbe2023generalization} or scratchpads \citep{anil2022exploring}. We firmly believe that an important trait of reasoning is robustness with respect to prompt quality---so long as the prompt unambiguously specifies the problem---and hence deliberately do not explore prompt modification approaches here; only randomised positions \citep{randomizedpe} are leveraged out of the works above in our model.

\paragraph{Tool use and multimodality} 

Another way to obtain robust generalisation performance is to leverage a hard-coded algorithm (also known as a \emph{tool}) by teaching an LLM to invoke its API \citep{schick2023toolformer}. Arguably, most of the major successes of reasoning with LLMs \citep{leblond2023ac2,romera2023mathematical,trinh2024solving} can primarily be attributed to an LLM's clever usage of a tool rather than the LLM itself, as a tool will by definition not have issues in generalising to diverse inputs.

Since our aim is to directly evaluate reasoning capabilities of LLMs, we explicitly do not permit tool use in our baselines. That being said, we envision the pre-trained NAR as a \emph{modulator} for the Transformer's embeddings which is more robust to OOD noise. Hence, we may observe the NAR as an \emph{``internal tool''}: rather than using raw tokens, the Transformer and NAR can communicate using their embeddings, breaking the associated algorithmic bottlenecks \citep{deac2021neural,ong2023probing}.

How to actually realise this communication and embedding exchange? For this, we turn to \emph{multimodal LLMs} \citep{jaegle2021perceiver} for inspiration, since we need to integrate signals coming from two different representations of algorithmic problems (text and graph). Specifically, our exchange operator is directly inspired by vision language models (VLMs) and the cross-attention operator used in Flamingo \citep{flamingo}, which offered a principled way of fusing information from text and image modalities.

\section{TransNAR: Augmenting Transformers with a pre-trained GNN-based NAR}
This section describes our hybrid TransNAR architecture (refer to Figure \ref{fig:architecture}). TransNAR accepts a dual input consisting of a textual algorithmic problem specification (of $T$ tokens) and its corresponding CLRS-30-specific graph representation (of $N$ nodes) and outputs a textual response to the problem. We can assume that, once encoded, the textual input is stored in ${\bf T}\in\mathbb{R}^{T\times k}$, and the graph input is stored in ${\bf G}\in\mathbb{R}^{N\times l}$. Note that, for simplifying the equations to follow, we make an assumption that all of the information relevant to the graph version of the problem is stored in the nodes---which is often not true in CLRS-30 (there may be edge- and graph-level inputs as well) but it doesn't change the underlying dataflow presented below.

The forward pass of TransNAR unfolds as follows. First, we properly initialise the inputs by setting ${\bf T}^{(0)} = {\bf T}$ and ${\bf G}^{(0)} = {\bf G}$. Next, to compute the representation of a step $(t+1)$, the text (token) representations are fed to the current layer of the Transformer \citep{vaswani2017attention}:
\begin{equation}
    {\mathbf{\Theta}}^{(t+1)} = \mathrm{FFN}\left(\mathrm{softmax}\left(\frac{({\bf T}^{(t)}{\bf Q}_t)^\top{\bf T}^{(t)}{\bf K}_t}{\sqrt{d_k}}\right){\bf T}^{(t)}{\bf V}_t\right)
\end{equation}
where ${\bf Q}_t, {\bf K}_t\in\mathbb{R}^{k\times d_k}, {\bf V}_t\in\mathbb{R}^{k\times k}$ are the key, query and value transformations, respectively, and $\mathrm{FFN}$ is a feedforward network. In a similar manner, the graph representations are fed to the NAR layer, implementing e.g. a standard max-MPNN \citep{velivckovic2019neural}:
\begin{equation}\label{eqn:mpnn}
    {\bf g}^{(t+1)}_u = \phi\left({\bf g}^{(t)}_u, \max_{1\leq v\leq N}\psi\left({\bf g}^{(t)}_u, {\bf g}^{(t)}_v\right)\right)
\end{equation}
where $\psi, \phi : \mathbb{R}^k\times\mathbb{R}^k\rightarrow\mathbb{R}^k$ are learnable \emph{message} and \emph{update} functions, respectively, and $\max$ is the elementwise-max aggregation. Note that Equation \ref{eqn:mpnn} only provides pairwise interactions between nodes for brevity---in reality, our NAR is a Triplet-GMPNN \citep{generalist}, which also contains triplet interactions and a gating mechanism. Further, note that there is no timestep index on the learnable parts of the NAR---at each step, a \emph{shared} function is applied. This aligns well with the iterative, repeated nature of algorithmic computation on graphs.

Once both streams have prepared their representations, $\mathbf{\Theta}^{(t+1)}$ and $\mathbf{G}^{(t+1)}$, the node embeddings in the graph condition the Transformer's token embeddings to produce the final outcome of the TransNAR block in the Transformer stream, inspired by Flamingo \citep{flamingo}:
\begin{equation}\label{eq:cross}
    {\bf T}^{(t+1)} = \mathrm{FFN}\left(\mathrm{softmax}\left(\frac{({\mathbf{\Theta}}^{(t)}{\bf Q}^\times_t)^\top{\bf G}^{(t)}{\bf K}^\times_t}{\sqrt{d_k}}\right){\bf G}^{(t)}{\bf V}^\times_t\right)
\end{equation}
where ${\bf Q}_t^\times, {\bf K}_t^\times\in\mathbb{R}^{k\times d_k}, {\bf V}_t^\times\in\mathbb{R}^{k\times k}$ are the key, query and value transformations of the cross-attention, respectively. No additional transformations are performed on ${\bf G}^{(t+1)}$ before concluding this layer. 

This process repeats until the final, $N_l$-th layer, when the final text output is read out from ${\bf T}^{(N_l)}$. The final output is converted into token logits by a prediction head produced by the final layer, which we supervise by means of a standard next-token prediction objective.

Prior to the start of TransNAR fine-tuning, we pre-train the NAR to robustly execute the thirty algorithms spanned by CLRS-30 \citep{clrs30}, in a manner similar to \citet{generalist}. Such procedures are known to yield out-of-distribution generalisation at up-to-$4\times$ larger inputs in graph space. The parameters of the NAR are generally kept \emph{frozen} during fine-tuning, as additional gradients would eliminate the model's original robustness properties. This is also, similarly, the reason why no cross-attention is performed by the graph embeddings. The LLM itself may be pre-trained over large-scale datasets \citep{chinchilla}, to establish its general language priors, though we recover the same experimental findings even if the LM is randomly initialised at the onset.
\section{Experiments}

\begin{table*}
\begin{center}
\begin{tabular}{l|l|c}
\toprule
              Algorithm &  \textcolor{mygreen}{Input} \& \textcolor{red}{Target} & \#tokens \\
\midrule
Articulation points &          \textcolor{mygreen}{\texttt{articulation\_points:}} & 63\\
& \textcolor{mygreen}{\texttt{A: [[0 0 0 0], [0 1 0 0], [0 0 0 0], [0 0 0 1]]}}\\
& \textcolor{mygreen}{\texttt{is\_cut:}} \textcolor{red}{\texttt{[0 0 0 0]}}\\
      Binary search &                          \textcolor{mygreen}{\texttt{binary\_search:}} & 49\\
      & \textcolor{mygreen}{\texttt{key: [0.011 0.029 0.635 0.719], target: 0.122}}\\
      & \textcolor{mygreen}{\texttt{return:}} \textcolor{red}{\texttt{2}} \\
     Insertion sort &                  \textcolor{mygreen}{\texttt{insertion\_sort:}} & 65\\ 
     & \textcolor{mygreen}{\texttt{key: [0.561 0.081 0.892 0.565]}}\\
     & \textcolor{mygreen}{\texttt{pred:}} \textcolor{red}{\texttt{[0.081 0.561 0.565 0.892]}} \\
       Jarvis' march &        \textcolor{mygreen}{\texttt{jarvis\_march:}}& 75\\
      &  \textcolor{mygreen}{\texttt{x: [-1.22 -1.05 0.331 -1.55], y: [-1.48 1.39 0.899 0.1]}}\\
      &  \textcolor{mygreen}{\texttt{in\_hull:}} \textcolor{red}{\texttt{[1 1 1 1]}}\\
        KMP Matcher &                                         \textcolor{mygreen}{\texttt{kmp\_matcher:}} & 37\\ 
        & \textcolor{mygreen}{\texttt{string: [0 0 0 1], key: [3 3 2 3]}}\\
        & \textcolor{mygreen}{\texttt{match:}} \textcolor{red}{\texttt{0}}\\
 Matrix Chain Order & \textcolor{mygreen}{\texttt{matrix\_chain\_order:}} & 77\\
& \textcolor{mygreen}{\texttt{p: [0.461 0.957 0.462 0.42]}}\\
& \textcolor{mygreen}{\texttt{s:}} \textcolor{red}{\texttt{[[0 0 0 0], [0 0 1 2], [0 0 0 2], [0 0 0 0]]}} \\
    Task Scheduling &                 \textcolor{mygreen}{\texttt{task\_scheduling:}} & 61\\ 
    & \textcolor{mygreen}{\texttt{d: [2 3 3 4], w: [0.042 0.875 0.954 0.761]}}\\
    & \textcolor{mygreen}{\texttt{selected:}} \textcolor{red}{\texttt{[0 1 1 1]}} \\
\bottomrule
\end{tabular}

\label{clrs-samples}
\caption{\textbf{Samples from CLRS-Text for various algorithmic tasks of problem size 4.} Input and target parts of the examples are clearly specified. Note that variability is limited at size 4, meaning that some algorithms may have trivial answers for the given inputs. Such effects tend to quickly disappear with scaling the problem size.}
\end{center}
\end{table*}

\begin{figure*}[ht]
\begin{center}
\includegraphics[width=\linewidth]{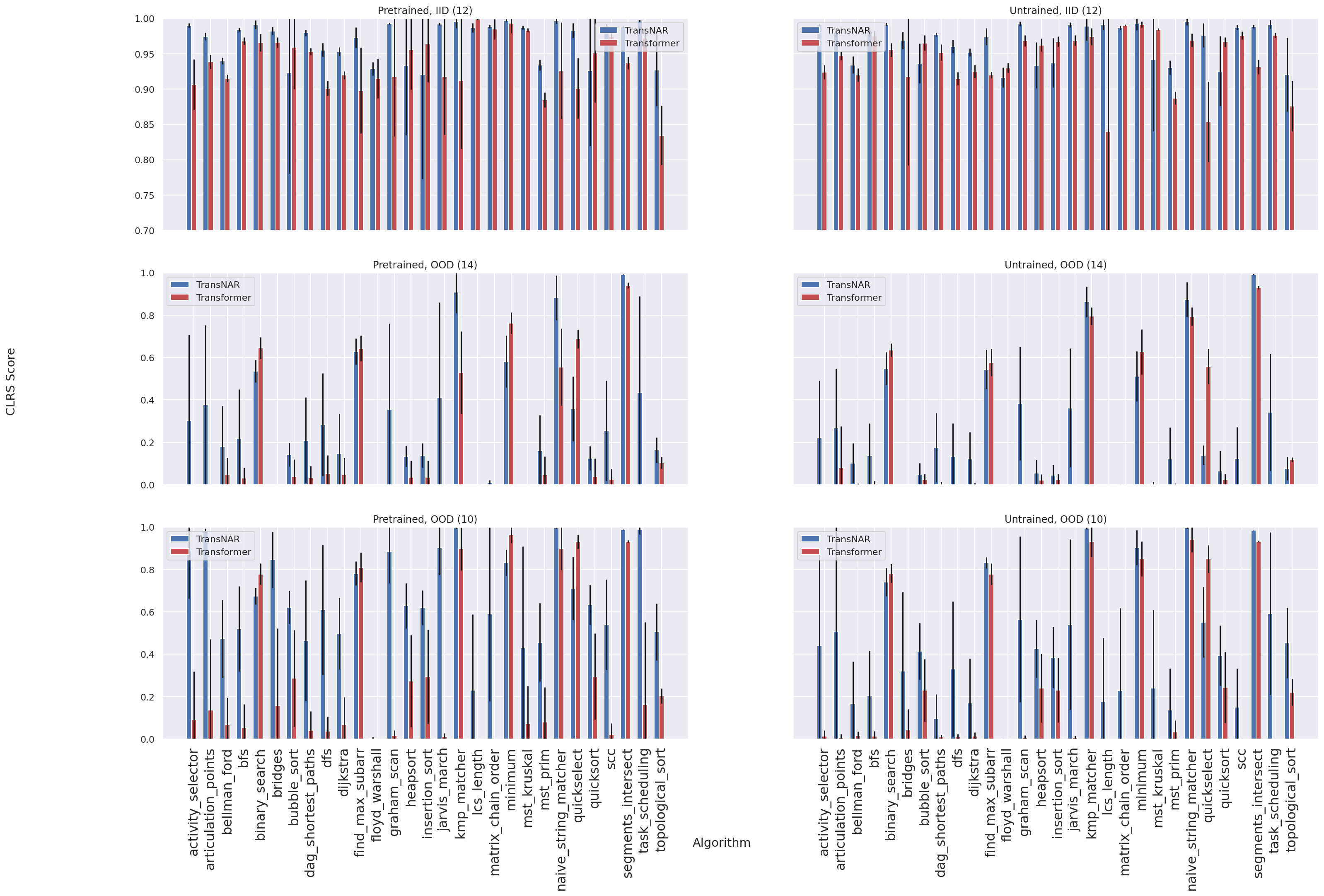}
\caption{\textbf{TransNAR significantly outperforms the baseline Transformer.} We compare TransNAR to its corresponding Transformer baseline on various algorithms and for various input sizes: $12$ is the largest size in-distribution. The other two sizes tested---$10$ and $14$---are out-of-distribution, with the former testing interpolation and the latter extrapolation. Note that in-distribution generalisation is much easier for Transformers, and as such, we have modified the $y$-axis for this setting only to the $[0.7, 1.0]$ range. It is evident that, on most algorithmic tasks of interest, the TransNAR is capable of outperforming its baseline Transformer. Additionally, we see that this advantage is consistent across both training regimes: initial training and finetuning. The metric used is the CLRS score. Each model was trained with 4 random seeds. Error bars indicate $\pm 1$ standard deviation.}
\end{center}
\label{clrs-score}
\end{figure*}

In our experimentation, we will demonstrate that the recipe offered by TransNAR admits significant benefits to out-of-distribution reasoning in language model architectures. In this section we provide details of our experimental setup.

\textbf{Transformer architecture and initialisation.} We use a decoder-only, 6 layers, transformer model from the Chinchilla family \citep{chinchilla} pretrained on MassiveText \cite{rae2022scaling}. In particular we use a model of $70$ million parameters with a context size $2,048$. To showcase the suitability of our approach regardless of the starting point of training, we run two ablative variants. In the first, the Transformer weights are initialised with the outcome of the pre-training---emulating a \emph{fine-tuning} scenario---and in the second, we use a fully random initialisation. In our figures and tables of results that follow, we will refer to these two setups as \textit{``Pretrained''} and \textit{``Untrained''}.

\textbf{Randomized positional encoding.} Previous work has emphasised the significant relevance of \emph{randomised} positional embeddings in Transformers, especially for enabling more robust reasoning \cite{randomizedpe}. Corresponding to previous studies on the generalization capabilities of language models, randomised positional embeddings have indeed led to significant gains on both our baselines and TransNAR, allowing more interesting reasoning behaviour to emerge in both. As such, all our experiments in this paper will use randomised positional embeddings. We provide more details in Appendix \ref{randomized-pos}.

\textbf{Pre-training the NAR.}
Following \citet{generalist}, we pre-train a multi-task MPNN-based NAR on input problem sizes of up to 16, from the CLRS-30 benchmark \citep{clrs30}. Owing to its graph structure formulation, such NARs are capable of significant OOD generalisation---sometimes staying competitive on graphs that are $4\times$ the size. We will attempt to utilise such models through TransNAR, to convey this rich representational knowledge into text. 

\textbf{Combining cross-attention contributions from nodes and edges.}
The NAR pre-trained by the method presented in \citet{generalist} produces both node and edge latent representations, and we cross-attend to both of them, as they may contain complementary useful information. To cross-attend over the edge features, ${\bf E}^{(t)}\in\mathbb{R}^{N\times N\times k}$, we apply Equation \ref{eq:cross} one more time (with $\mathbf{\Theta}^{(t)}$ cross-attending over ${\bf E}^{(t)}$), with the caveat that we need to flatten the first and second axis of ${\bf E}$ into one, to make sure the dimensionalities match. We combine the cross-attention contribution from the node and edge embeddings provided by the pre-trained NAR by concatenation, followed by the application of a linear layer. We have attempted to use other reduction schemes such as summing the vectors, or applying a 2-layer MLP. We have also attempted different preprocessing schemes such as orthogonalising the contributions using the Gram‐Schmidt process to ensure their algebraic complementarity before combining them. However, none of these variations have brought improvements over our original approach.

\textbf{Datasets.}
We use the CLRS-Text benchmark \cite{markeeva2024clrstext}, the text version of the CLRS-30 benchmark \cite{clrs30}.
Table 1 showcases several samples from this dataset, along with their input size and number of tokens. Note that the textual representation is directly derived from the graph-based CLRS-30 in a deterministic manner, so the two datasets convey exactly the same information. However, due to the tokenised representation, there are stringent limitations on how large of a problem size we can evaluate on without running out of context length for the Chinchilla models.

Accordingly, we train our algorithms on smaller problem sizes---$[4,8]$ and $12$, and evaluate on problem sizes $10$ (\emph{out-of-distribution---interpolation}), $12$ (\emph{in-distribution}), $14$ (\emph{out-of-distribution---extrapolation}).

It is worth noting that CLRS-Text is among the most challenging long-range reasoning tasks for language models, compared to the present evaluation landscape---a clear step-up in complexity from grade school math, mainly because it allows for explicitly controlling for out-of-distribution generalisation. Yet, there exists a clear polynomial-time-algorithmic description for each of them, meaning that they can be explained in relatively little parameters---certainly way less than a typical large language model of today!

The dataset comprises $10,000$ samples per algorithm per input size, making up a total of $2,400,000$ data points, split as per above into $70\%$ for training and $30\%$ for validation. 

\paragraph{Training details.} We train all models over seven epochs of the training data with a batch size of $256$ and employ an Adam optimizer \cite{adam} with a learning rate of $10^{-4}$. We apply randomized positional encoding with a maximal length of $8,192$ on top of Rotary Positional Encoding (RoPE) used in the base Chinchilla transformer \cite{chinchilla}. As previously mentioned, for all TransNAR models, we keep the NAR frozen during training.

\begin{figure*}[ht]
\label{shape-score}
\begin{center}
\includegraphics[width=\linewidth]{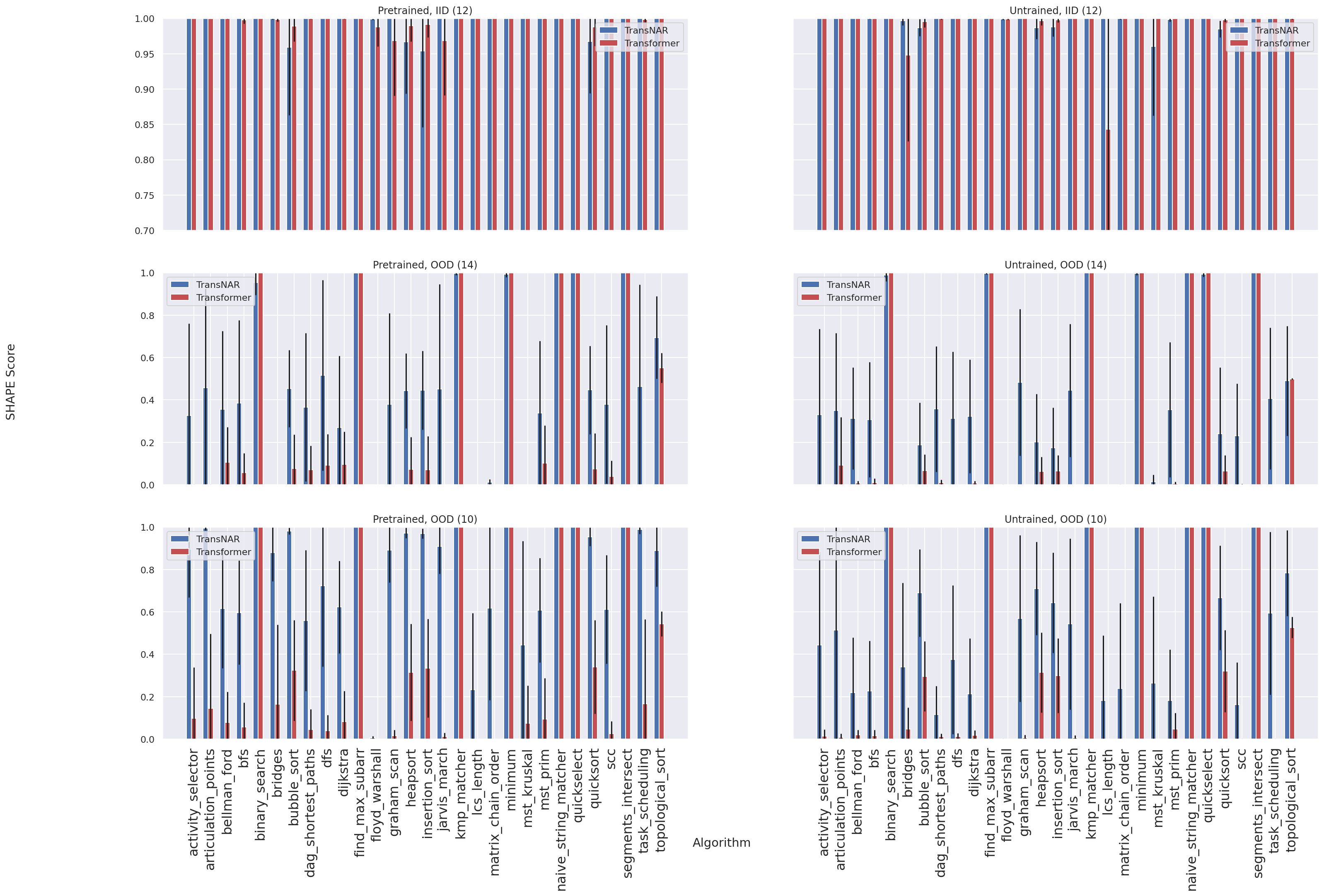}
\caption{\textbf{Shape Score:} The TransNAR significantly outperforms its baseline in terms of producing correct shapes. This score sheds light on an obvious failure model of regular Transformers out-of-distribution: they fail to capture the seemingly trivial dependency between input size and output size, and so irrespective of the complexity of the algorithm itself. The TransNAR model manages to considerably alleviate this problem (with many emerging gains), albeit, these gains do not always lead to perfect scores, implying a fruitful direciton for future research.}
\end{center}
\end{figure*}

\textbf{Evaluation metrics.} We refrain from computing the accuracy of each model using exact string matching, on the grounds that this metric does not provide insights as to the causes of failure on a particular datapoint, and more critically, it fails to capture how close to correctness a given model output is (as observed by \citet{clrs30}). Instead, we evaluate the performance of each model according to three metrics measuring capabilities of increasing complexity over the generated text:
\begin{enumerate}
    \item The \textit{shape score}: a binary-valued metric capturing whether the output has the right shape. For example, if we consider a sorting task, the output should have exactly the same number of elements as the input. Similarly, if the output is a matrix, we ensure its shape is consistent with both the input and the task.
    \item The \textit{parse score}: a binary-valued metric capturing whether the output is free from any illicit characters, for example, considering again a sorting task on a list of numbers, the output shouldn't contain any letters of the alphabet.
    \item The \textit{CLRS score}: The percentage of elements in the output that match the ground truth answer. This score is the one traditionally used in CLRS-30 \cite{clrs30, generalist}, hence its name. Note that we automatically assign a CLRS score of 0 if the shape score is 0, as there is no clear correspondence between output indices.
\end{enumerate}
These multi-faceted scores are explicitly designed to capture the various failure modes of LLMs when learning to reason over text inputs: they may overly specialise to the training problem sizes (leading to incorrect shapes at test time), fail to cope with unseen number combinations (leading to incorrect parsing), and of course, produce incorrect or inconsistent outputs, captured by the CLRS score.
\subsection{Results}
We summarize our findings in Figure 4 (for CLRS score). Our results show that our TransNAR significantly outperforms the baseline Transformer overall, and on most individual algorithms, both in- and out-of-distribution. In particular, we see that our approach not only enhances existing out-of-distribution generalisation capabilities, but also causes the emergence of these capabilities when there was a complete lack thereof---reflected in the figure by zero or near-zero performance of the baseline \citep{wei2022emergent}. 

The analysis of shape score (Figure 5) provides an additional way to shed light on why TransNAR performed as well as it did. Recall, first, that CLRS score is necessarily zero if shapes do not match. Observing the shape scores achieved, it appears that grounding Transformer outputs in NAR embeddings significantly increases the proportion of inputs for which a Transformer will produce an output of the correct shape---indicating that this is one very specific failure mode that TransNAR helps alleviate.

We note, however, that there remain a few algorithms for which TransNAR is not able to outperform the baseline. A closer look at the results indicates that such tasks (Binary Search, Find Maximum Subarray, Minimum, and Quickselect) all involve an element of \emph{searching} for a particular index in an input list. This hints at a unified failure mode: as these failures persist both when interpolating and extrapolating, the model as implemented is not able to generalise to novel \emph{index boundaries} unseen in the training data. We therefore suspect that the use of \emph{index hints}---as already demonstrated by \citet{zhou2023algorithms}---is a promising avenue for ameliorating this behaviour. Alternatively, it might be the case that the final NAR-computed hidden states are harder to decode by the cross-attention layers in a generalisable way, and therefore might require either giving an additional capacity to the cross-attention and/or performing a more \textit{progressive} decoding in that: instead of having all cross-attention layers decoding from the final NAR-computed hidden states, s, we could have early cross-attention layers decode from hidden states coming from earlier message passing steps, and later cross-attention layers decode from the later message passing steps.

Lastly, we provide parse scores in Appendix \ref{parse-and-shape-scores}---omitting them from the main text because, in most cases, parsing can be done at full accuracy.

\subsection{Limitations}
While our approach demonstrates favourable average performance under all out-of-distribution regimes we have evaluated, we highlight that TransNAR requires access to both textual and graph-representation inputs to be efficiently trainable and usable. While this limits TransNAR to cases where a particular ground-truth executor or simulator (or prior belief about one) is available, now that we know that TransNAR-like ideas are beneficial, future research can enable the deployment of such ideas into purely unimodal Transformers. For example, lifting the need for a second data stream can be done by \emph{distilling} the knowledge acquired by the trained TransNAR model into a vanilla Transformer model.
\section{Conclusions}
We presented a Transformer-NAR hybrid architecture: a language model that combines the language understanding skills of a Transformer with the robust algorithmic reasoning capabilities of a pre-trained graph neural network-based neural algorithmic reasoner, to solve algorithmic tasks specified in natural language. We have demonstrated the superiority of our model over its Transformer-only counterpart on the CLRS-text benchmark, in the in-distribution, and more importantly, in two out-of-distribution regimes, with respect to the input problem size. We hope that future work will draw on our results and insights shared here, and further investigate expansions of interest, notably, datasets with more ambiguous problem specifications (as often encountered in the real world), and for which their corresponding equivalent solver-ready symbolic inputs are not given in advance.

{
    \small
    \bibliographystyle{ieeenat_fullname}
    \bibliography{main}
}

\section{Effect of Randomized Positional Encoding}
\label{randomized-pos}
Using randomized positional encoding has benefitted both our model and the baseline. In particular, combining them with NAR hiddens led to improvements OOD, most prevalently in the interpoloation regime (at length 10), but also, to some extent, in the extrapoloation regime (at length 14). One result we found interesting, was that before instating randomized positional encoding, the OOD performance of our hybrid models was limited (in fact thresholded) by the performance of the base LLM. Concretely, if the base LLM achieved near-zero performance, the hybrid architecture would fatally share the same fate. We can see that this is no longer the case: if the base LLM uses randomized positional encoding, even if its performance is near-zero, that of the hybrid architecture can still be reasonably good. This is illustrated in the second column of the figure 4, for example on the Graham Scan, Jarvis March, MST Prim algorithms.

\section{Parse Scores}
\label{parse-and-shape-scores}
\begin{figure*}[ht]
\begin{center}
\includegraphics[width=\linewidth]{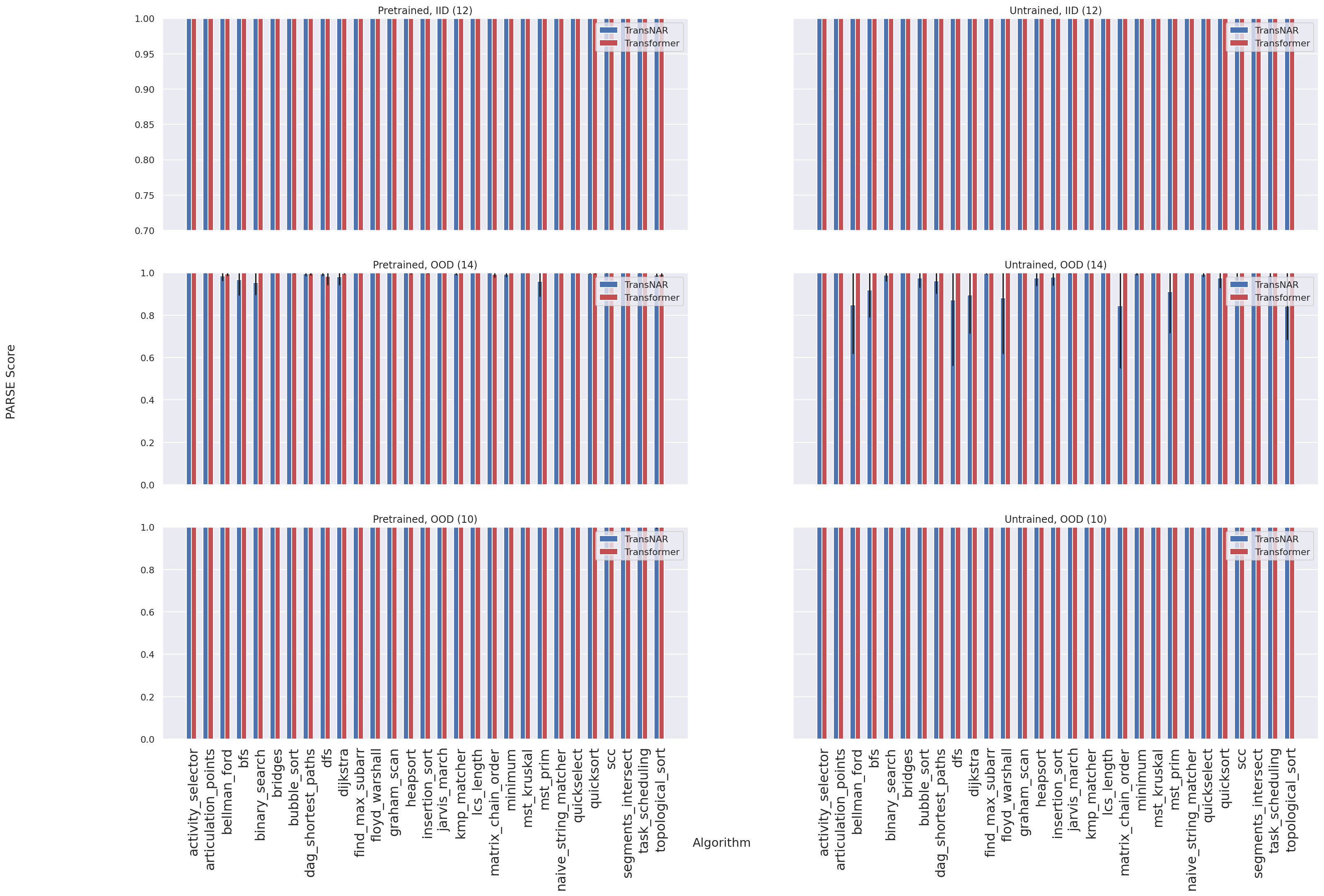}
\label{parse-score}
\caption{\textbf{Parse Score:} We can see that for a few algorithms, the TransNAR architecture falls behind the baseline in the extrapolation regime likely due to an unsufficient capacity of the cross-attention in charge of decoding from the NAR's outputs.}
\end{center}
\end{figure*}


\end{document}